\documentclass[preprint,12pt]{elsarticle}
\usepackage{hyperref}
\hypersetup{
colorlinks=true,
linkcolor=blue,
filecolor=blue,
urlcolor=blue,
pdftitle={Overleaf Example},
pdfpagemode=FullScreen,
citecolor=blue,
}
\usepackage{amsmath}
\usepackage{listings}
\usepackage{xcolor}
\usepackage{soul}

\definecolor{codegreen}{rgb}{0,0.6,0}
\definecolor{codegray}{rgb}{0.5,0.5,0.5}
\definecolor{codepurple}{rgb}{0.58,0,0.82}
\definecolor{backcolour}{rgb}{0.95,0.95,0.92}

\lstdefinestyle{mystyle}{
    backgroundcolor=\color{backcolour},   
    commentstyle=\color{codegreen},
    keywordstyle=\color{magenta},
    numberstyle=\tiny\color{codegray},
    stringstyle=\color{codepurple},
    basicstyle=\ttfamily\footnotesize,
    breakatwhitespace=false,         
    breaklines=true,                 
    captionpos=b,                    
    keepspaces=true,                 
    numbers=left,                    
    numbersep=5pt,                  
    showspaces=false,                
    showstringspaces=false,
    showtabs=false,                  
    tabsize=2
}

\usepackage{graphicx}
\usepackage{amssymb}

\newcounter{bla}

\journal{arXiv}

\begin{document}

\begin{frontmatter}

\title{ALT: A Python Package for Lightweight Feature Representation in Time Series Classification}

\author[a,b]{Bal\'azs P. Halmos}
\author[c,b]{Bal\'azs Haj\'os}
\author[a,b]{Vince \'A. Moln\'ar}
\author[d]{Marcell T. Kurbucz\corref{author}}
\author[b,e]{Antal Jakov\'ac}

\cortext[author] {Corresponding author.\\\textit{E-mail address:} m.kurbucz@ucl.ac.uk}

\address[a]{Faculty of Engineering and Natural Sciences, Tampere University, Kalevantie 4, Tampere, 33100, Finland}
\address[b]{Department of Computational Sciences, Wigner Research Centre for Physics, 29-33 Konkoly-Thege Miklós Street, Budapest, 1121, Hungary}
\address[c]{Faculty of Science, Eötvös Loránd University, 1/A Pázmány Péter Walkway, Budapest, 1117, Hungary}
\address[d]{Institute for Global Prosperity, The Bartlett, University College London, 149 Tottenham Court Road, London, W1T 7NF, United Kingdom}
\address[e]{Department of Statistics, Corvinus University of Budapest, 8 Fővám Square, Budapest, 1093, Hungary}

\begin{abstract}
We introduce \texttt{ALT}, an open-source Python package created for efficient and accurate time series classification (TSC). The package implements the adaptive law-based transformation (ALT) algorithm, which transforms raw time series data into a linearly separable feature space using variable-length shifted time windows. This adaptive approach enhances its predecessor, the linear law-based transformation (LLT), by effectively capturing patterns of varying temporal scales. The software is implemented for scalability, interpretability, and ease of use, achieving state-of-the-art performance with minimal computational overhead. Extensive benchmarking on real-world datasets demonstrates the utility of \texttt{ALT} for diverse TSC tasks in physics and related domains. \\

\noindent \textbf{PROGRAM SUMMARY}

\begin{small}
\noindent
{\em Program Title:} ALT                                          \\
{\em CPC Library link to program files:} (to be added by Technical Editor) \\
{\em Developer's repository link:} \href{https://github.com/Datacompintensive/ALT}{https://github.com/Datacompintensive/ALT} \\
{\em Licensing provisions(please choose one):} GPLv3  \\
{\em Programming language:} Python (3.9+)                                 \\ \\
{\em Nature of problem:} \\
Efficient and interpretable classification of time series data remains a fundamental challenge in computational physics. Time series generated from physical simulations, experimental data acquisition, and sensor networks often exhibit variable temporal patterns, noise, and complex dependencies. Existing approaches, such as deep learning and statistical techniques, tend to be computationally expensive, require extensive parameter tuning, or lack interpretability. Consequently, there is a need for a robust, scalable, and transparent method specifically tailored to physics-related time series classification. \\ \\
{\em Solution method:} \\
The software implements the adaptive law-based transformation (ALT) method. By extracting governing laws from time series data using time-delay embedding and variable-length shifted windows, \texttt{ALT} transforms features into a space optimized for linear separability. The adaptive windowing technique captures patterns across multiple temporal scales, enhancing classification accuracy without significant computational cost. Written in Python, the software efficiently processes time series data, enabling classification with traditional machine learning algorithms. \\ \\
{\em Additional comments including restrictions and unusual features:} \\ 
\begin{sloppypar} 
\noindent
Restrictions: The software is optimized for uni- and multivariate time series datasets with moderate dimensionality. For extremely large datasets, computational efficiency may depend on hardware capabilities and preprocessing strategies.
Unusual Features: \texttt{ALT} creates interpretable features with the transformation. The method is robust against noise and adaptable to varying input resolutions, making it particularly well-suited for physics-related applications where transparency is crucial. \\
\end{sloppypar}
\end{small}
   \end{abstract}


\end{frontmatter}

\section{Introduction}
\label{sec:introduction}

\noindent
Time series classification (TSC) is central to computational physics, where temporal data streams arise from experimental measurements, simulations, and sensor networks. Traditional TSC methods, including feature-based techniques~\cite{fulcher2014highly}, distance-based metrics~\cite{senin2008dynamic}, and deep learning models~\cite{fawaz2019deep}, each offer unique advantages but also face distinct limitations. In particular, challenges persist due to the high variability, noise, and complexity inherent in physical time series data.

Feature-based methods focus on extracting meaningful representations from time series before applying classification algorithms. These representations can include statistical descriptors~\cite{fulcher2014highly}, spectral transformations such as the discrete Fourier transform (DFT) or discrete wavelet transform (DWT)~\cite{agrawal1993efficient}, and model-based features derived from autoregressive integrated moving average (ARIMA) models~\cite{box2015time}. Shapelet-based techniques, which identify discriminative subsequences~\cite{ye2009time}, excel at capturing localized variations but may struggle with multi-scale patterns and the computational overhead of long time series. Common classification approaches for these representations include logistic regression, random forests, and support vector machines (SVMs).

Distance-based methods compare entire time series directly, avoiding explicit transformations into feature vectors. Dynamic time warping (DTW)~\cite{senin2008dynamic}, a widely used distance-based technique, effectively aligns time series by accounting for local temporal distortions, making it robust across many applications. However, these methods often suffer from high computational complexity for large datasets and provide limited interpretability due to the absence of intermediate features. Despite these drawbacks, their simplicity and alignment capabilities make them valuable tools for TSC.

Deep learning–based approaches have emerged as powerful alternatives, leveraging neural networks to automatically learn hierarchical features from raw time series data. Convolutional neural networks (CNNs) are particularly adept at detecting local temporal patterns, while recurrent neural networks (RNNs) excel at capturing sequential dependencies, including long-term patterns~\cite{fawaz2019deep, karim2019multivariate, zheng2014time}. Although these methods can achieve high classification accuracy, they often require large labeled datasets, significant computational resources, and careful hyperparameter tuning. Their lack of interpretability can also pose challenges in domains requiring explainable decisions.

Our earlier method, linear law-based transformation (LLT) \citep{jakovac2022reconstruction, jakovac2021time, kurbucz2022facilitating, KURBUCZ2024101623}, combines feature- and distance-based strategies by utilizing time-delay embedding and spectral decomposition to enhance classification efficiency. However, its reliance on fixed-length time windows limits adaptability, restricting its ability to handle datasets with varying temporal structures.

To overcome this limitation, we developed the adaptive law-based transformation (ALT) \citep{kurbucz2025adaptive}, which extends LLT by introducing variable-length shifted time windows, enabling better representation of multi-scale temporal patterns. ALT preserves LLT’s core principle of mapping time series data into a linearly separable feature space, but with enhanced flexibility and robustness.

This article presents the software package \texttt{ALT}, providing an efficient implementation of the ALT method. By automating feature extraction and transformation, the software facilitates practical applications of ALT, making it a computationally efficient and interpretable alternative to existing TSC techniques.

We evaluated \texttt{ALT} using benchmark time series datasets and demonstrated~\cite{kurbucz2025adaptive} its superior performance compared to traditional methods, particularly for data with complex temporal dependencies. Its interpretability and efficiency make it well-suited for computational physics applications, as well as other scientific domains—from analyzing health-related time series data~\cite{wu2024invariant} to identifying patterns in finance~\cite{kurbucz2023predicting}.

The rest of the paper is organized as follows: Section \ref{sec:algorithm} describes the transformation algorithm, and Section \ref{sec:software-description} presents the software architecture, input parameters, and example code snippets. Section \ref{sec:illustrative-examples} illustrates the feature space transformation through practical examples. Finally, Section \ref{se:conclusions} concludes the article.

\section{Algorithm}
\label{sec:algorithm}

\noindent A general TSC problem can be defined as follows: The input data is represented as $x_t^{i,j}$, where $t \in \{1, 2, \dots, h\}$ denotes observation times, $i \in \{1, 2, \dots, \tau\}$ identifies instances, and $j \in \{1, 2, \dots, m\}$ corresponds to different input series for a given instance. The goal is to predict the class $y^i \in \{1, 2, \dots, c\}$ of instance $i$ based on the provided data. The proposed approach follows these steps:

\begin{enumerate}[noitemsep]

\item[\textbf{[A1]}] \textbf{Data Partitioning.} Randomly split the instances into learning ($Lr$), training ($Tr$), and test ($Te$) subsets while ensuring balanced class representation via stratified sampling.

\item[\textbf{[A2]}] \textbf{Sequence Extraction.} For each combination of $Lr$, $j$, and the parameter triplet $(r, l, k)$, extract sequences of length $r$ using time-shifted windows with a step size of $k$. Downsample these sequences to $2l-1$ points, where the predefined parameters satisfy $r \leq h$ and $(2l-2) \mid (r-1)$. The total number of sequences generated per configuration is $\lfloor (h - r + 1) / k \rfloor$.

\item[\textbf{[A3]}] \textbf{Shapelet Vector Computation.} Perform $l$-dimensional time-delay embedding \cite{takens1981dynamical} to construct a symmetric matrix $S$, where the sequence length is $2l-1$. Compute the spectral decomposition of $S$ and take the eigenvector corresponding to the smallest absolute eigenvalue as the shapelet vector $v$, satisfying $Sv \approx 0$. Unlike principal component analysis (PCA) \cite{gao2021human}, which emphasizes maximum variance, this step identifies low-variability dimensions to extract shapelets.

\item[\textbf{[A4]}] \textbf{Shapelet Matrix Formation.} Group shapelet vectors derived from the same $j$ and $(r, l, k)$ settings as column vectors in the shapelet matrix $P$. Further, partition $P$ according to class labels, forming $c$ class-specific subsets.

\item[\textbf{[A5]}] \textbf{Instance Transformation.} Transform each $Tr$ instance by embedding its time series into an $o \times l$ matrix $A$, where $o = \lfloor (h - sl + 1) / k \rfloor$ and $s = (r - 1) / (2l - 2)$, as follows:

\begin{equation}
    A = 
    \begin{pmatrix}
    x_1^{i, j} & x_{s+1}^{i, j} & \dots  & x_{(l-1)s+1}^{i, j} \\
    x_{k+1}^{i, j} & x_{k+s+1}^{i, j} & \dots & x_{k+(l-1)s+1}^{i, j} \\
    \vdots & \vdots & \ddots & \vdots \\
    x_{(o-1)k+1}^{i, j} & x_{(o-1)k+s+1}^{i, j} & \dots & x_{(o-1)k+(l-1)s+1}^{i, j} \\
    \end{pmatrix}.
\end{equation}

Multiply $A$ with the corresponding shapelet matrix $P$ to obtain $O = AP$. The shapelets in $P$ associated with different classes ``compete'' to transform the rows of $A$ towards near-null vectors.

\item[\textbf{[A6]}] \textbf{Feature Extraction.} Partition $O$ according to class-specific shapelet matrices and extract features using various statistical methods. For instance, compute the 5\(^\text{th}\) percentile of each row in $O$ and then take the average. This process yields a feature representation in an $m \times c \times n \times g$ space, where $n$ represents the number of extraction methods and $g$ is the number of $(r, l, k)$ configurations.

\item[\textbf{[B1]}] \textbf{Classifier Training and Validation.} Train a classifier (e.g., $K$-nearest neighbors) using the extracted features. Optimize hyperparameters via Bayesian optimization and evaluate performance using cross-validation. Report accuracy and computational efficiency.

\item[\textbf{[B2]}] \textbf{Testing and Benchmarking.} Apply the trained model to the test set $Te$, repeating steps [A5--A6] for transformation and classification. Assess accuracy, computational performance, and compare results against state-of-the-art models.

\end{enumerate}

\section{Software description}
\label{sec:software-description}

\subsection{Software architecture}

\begin{figure}[h]
    \centering
    \includegraphics[width=0.3\linewidth]{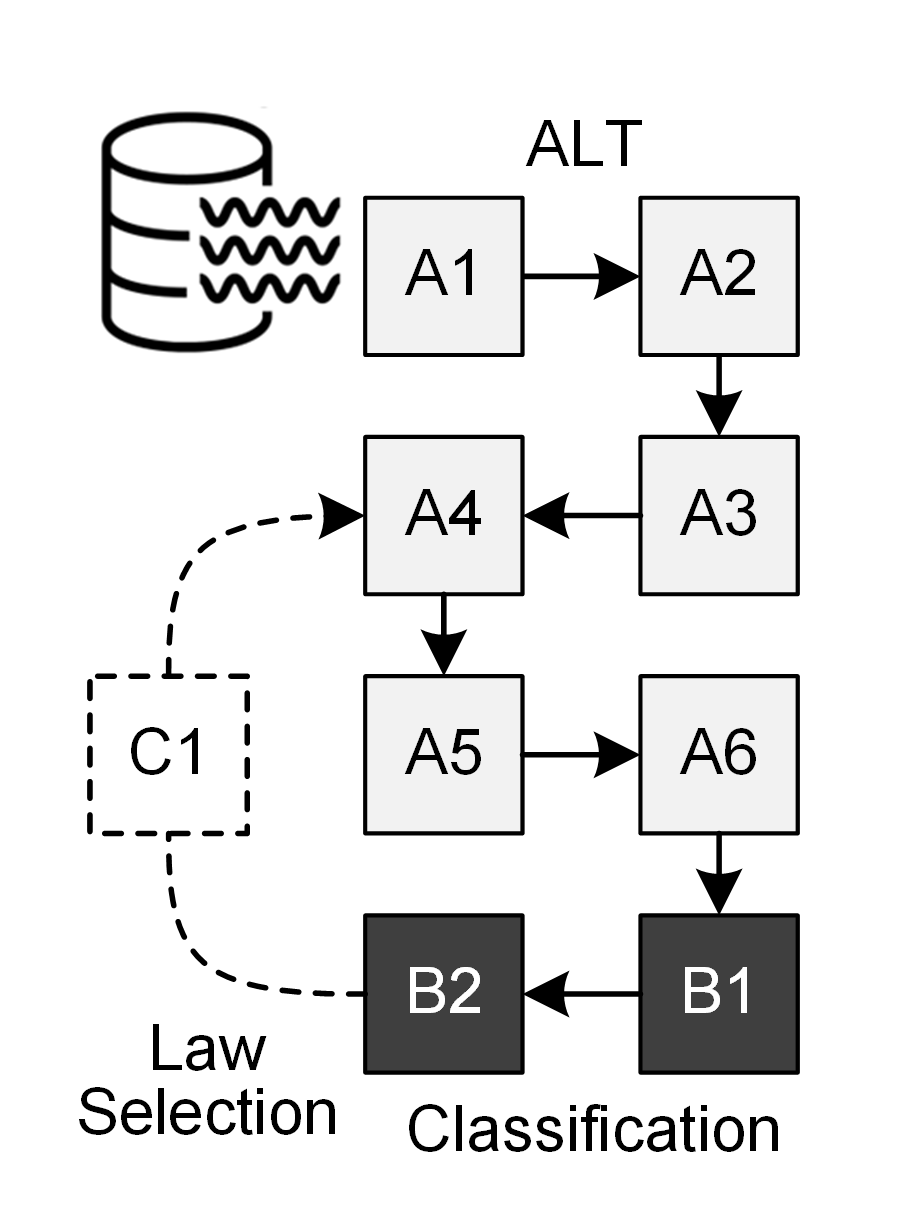}
    \caption{Software architecture. The blocks correspond to the points in Section \ref{sec:algorithm}. Original source: \cite{kurbucz2025adaptive}.}
    \label{fig:ALT}
\end{figure}

\noindent
Figure \ref{fig:ALT} illustrates the complete data transformation and classification pipeline.
The \texttt{ALT} package converts the data into easily separable features [A1–A6], after which traditional classifiers---such as those available in the Classification Learner App in MATLAB\footnote{More information is available at \url{https://www.mathworks.com/help/stats/classificationlearner-app.html} (retrieved: January 30, 2025).}---can be used for final classification [B1–B2]. Step C1 is planned for future implementation; see Section \ref{se:conclusions}.

\subsection{Software functionalities}

\noindent
The parameters of the \texttt{ALT} constructor:
\begin{itemize}
    \item \texttt{train\_set} (numpy.ndarray or torch.Tensor) The data from which the shapelets are generated for the transformation.
    \item \texttt{train\_classes} (numpy.ndarray or torch.Tensor) To which class do the instances belong in the \texttt{train\_set}.
    \item \texttt{train\_length = None} (numpy.ndarray or torch.Tensor) If the train instances have different length instances, it describes the length of each instance.
    \item \texttt{R = None} (list or int) The window size \(r\).
    \item \texttt{L = [5]} (list or int) The embedding dimension \(l\).
    \item \texttt{K = 1} (list or int) The window translation size \(k\).
    \item \texttt{device = "cpu"} On which device the code is run (``cuda'' is the GPU).
    
    Note: The length of \texttt{R, L} and \texttt{K} must be the same if given as lists (if given as int, that value will be used). 
\end{itemize}

The parameters of the \texttt{train} function:
\begin{itemize}
    \item \texttt{cleanup = False} (bool) Whether to remove the original (untransformed) data from the memory.
\end{itemize}

The parameters of the \texttt{save} function:
\begin{itemize}
    \item \texttt{save\_file\_name} (str) Saves the trained \(P\) matrices.
    
    Note: it can only be called once the model has been trained.
\end{itemize}

The parameters of the \texttt{ALT.load} function:
\begin{itemize}
    \item \texttt{load\_file\_name} (str) Loads the trained \(P\) matrices.
    
    Note: it does not read the original data, only the transformed.
\end{itemize}

The parameters of the \texttt{transform} function:
\begin{itemize}
    \item \texttt{z} (numpy.ndarray or torch.Tensor) An instance to be transformed.
    \item \texttt{extr\_methods} (list) A 2D list of the feature extraction methods in [A6]. For example, \texttt{extr\_methods = [["mean\_all"], ["4th\_moment", 0.05]]} calculates using two different extraction methods: the first calculates the mean of the \(O\) matrix, the second calculates the 5\(^\text{th}\) percentile of each row in \(O\) then calculates the kurtosis of the percentiles.  
\end{itemize}

The parameters of the \texttt{transform\_set} function:
\begin{itemize}
    \item \texttt{test\_set} (numpy.ndarray or torch.Tensor) The instances to be transformed.
    \item \texttt{extr\_methods} (list) A 2D list of the feature extraction methods.
    \item \texttt{test\_length = None} (numpy.ndarray or torch.Tensor) If the test instances have different length instances, it describes the length of each instance.
    \item \texttt{save\_file\_name = None} (str) The filename for saving the transformed features. Note: must be given if \texttt{save\_file\_mode} is given.
    \item \texttt{save\_file\_mode = None} (str) The mode in which the results are saved, can only be ``New\_file'', ``Append feature'', or ``Append instance''. Note: must be given if \texttt{save\_file\_name} is given.
    \item \texttt{test\_classes = None} (numpy.ndarray or torch.Tensor) The classes for the transformed set. Only used when saving, otherwise None.
\end{itemize}

 \subsection{Sample code snippets analysis (optional)}

\noindent
The \texttt{ALT} library can be imported as follows. Then the aeon library is used to import the data from the UCR \cite{UCRArchive2018} Time Series Classification Archive\footnote{These datasets are available at: \url{https://www.timeseriesclassification.com} (retrieved: January 30, 2025).}. Note that the data is shuffled in the original \textit{GunPoint}~\cite{ratanamahatana2005three} dataset, if that is not the case, an additional shuffling might be needed here. Also note that originally there are 200 instances in the \textit{GunPoint} database: 50 train and 150 test. From the train instances 10 are used to train the ALT method, and everything else is transformed. The remaining 40 train instances can be used to train a KNN or SVM classifier in MATLAB and the 150 test instances can be used to test it.

\begin{lstlisting}[language=Python, caption=Importing the library and the data]
# Imports
from ALT import *
from aeon.datasets import load_classification

# Import data from aeon
X, y = load_classification("GunPoint")
y = y.astype(np.int8)
# The data is shuffled
learn_set, transform_set = X[:10], X[10:]
learn_classes, transform_classes = y[:10], y[10:]
\end{lstlisting}

Then the parameters for the feature space transformation are defined: the time window used, the dimension of the laws, the extraction methods used, and the device, which tells the code to run on the GPU.

The \texttt{ALT} is initialized with the defined hyperparameters. Then the model is trained and the data is transformed.

\begin{lstlisting}[language=Python, caption=Transforming the data]
# Seting the parameters
R, L, K = 25, 4, 1
extr_methods = [["mean_all"], ["mean", 0.05]]
device = torch.device("cuda")

#Transform the data
alt = ALT(learn_set, learn_classes, R=R, L=L, K=K, device=device)
alt.train()
transformed_set = alt.transform_set(transform_set, extr_methods=extr_methods, test_classes=transform_classes, save_file_name="features.csv", save_file_mode="New file")
\end{lstlisting}

The features are saved to the features.csv file, which can be imported to MATLAB, or other similar programs for further classification.

\section{Illustrative examples}
\label{sec:illustrative-examples}

\noindent
This section presents simple examples of using the \texttt{ALT} package. The first example shows that after the transformation the data is linearly separable by forming clusters. The second example demonstrate the linear separability on another database. Finally, the effects of the different parameters (\(r,l,k\)) on the classification accuracy are detailed.

\subsection{\texttt{ALT} creates clusters of features}

\noindent
The example employs the \textit{BasicMotions} dataset, sourced from the UCR Time Series Classification Archive. The dataset aims to differentiate between four activities, walking, resting, running and badminton, using the data collected from the 3D accelerometer and the 3D gyroscope of a smart watch.

\begin{figure}[h]
    \centering
    \includegraphics[width=0.6\linewidth]{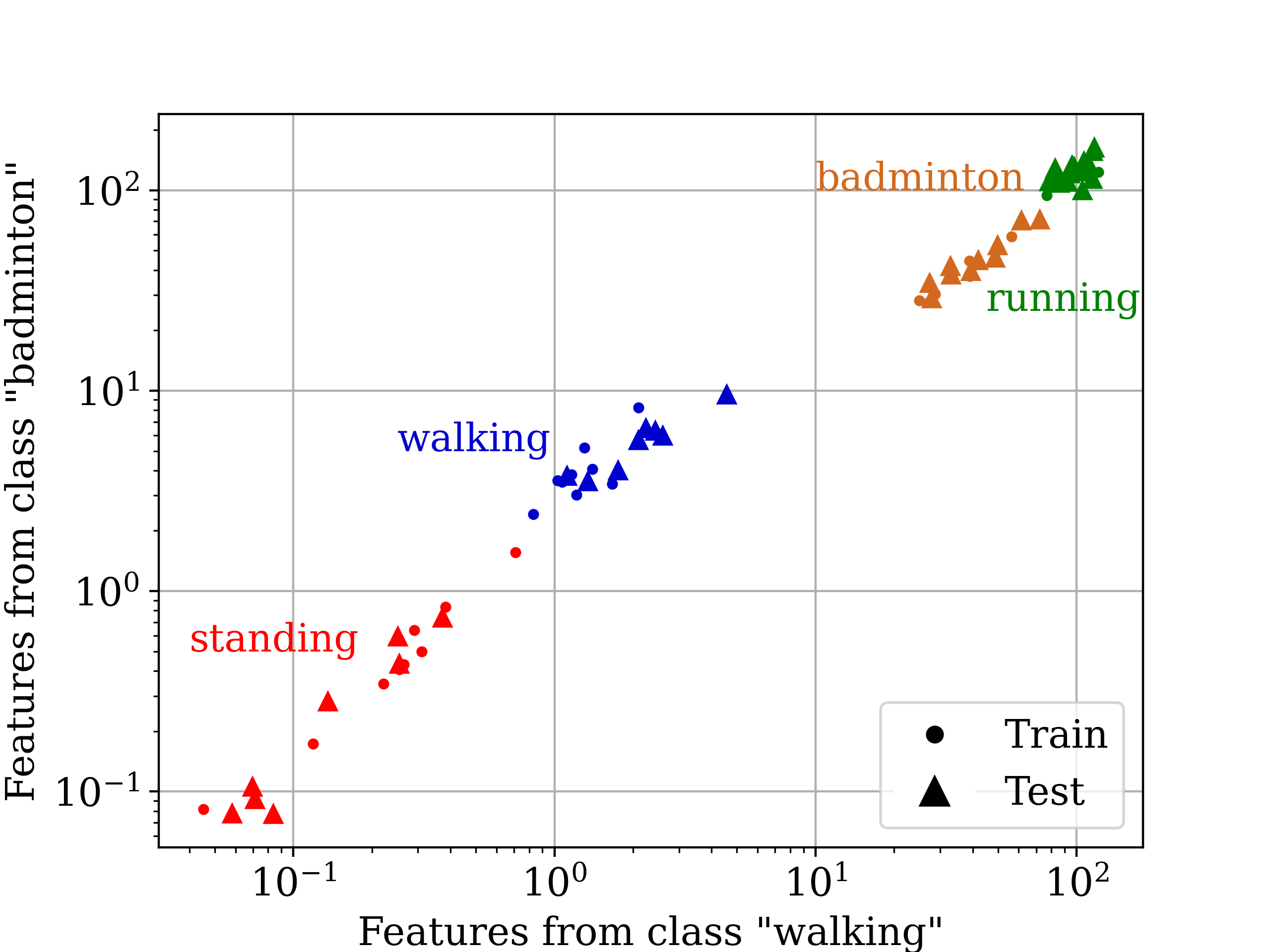}
    \caption{The transformation makes the data form clusters. The \textit{BasicMotions} dataset, \(r=53,\, l=27,\, k=1\), 2nd sensor, meanAll as the feature extraction method. The instances corresponding to different classes are marked with different colors, and can be separated from each other easily, the train instances are marked with dots, and the test instances with triangles.}
    \label{fig:BasicMotions_plot}
\end{figure}

After the transformation, the instances which belong to the same class form separable clusters, Figure \ref{fig:BasicMotions_plot}. The plot depicts two of the generated features: one was generated from the walking and the other from the badminton instances after calculating the mean of all the numbers in [A6] step. The parameters used were \(r=53,\, l=27,\, k=1\), and the 2nd sensor is shown.

After the transformation, even simple classification algorithms are able to separate the data with high accuracy. For instance, the Fine KNN (1 neighbor) classifier in the Classification Learner App in MATLAB is capable of classifying with 100\% validation and test accuracy only on these 2 features.

\subsection{\texttt{ALT} tries to create a linearly separable feature space}

\begin{figure}[h!]
    \centering
    \includegraphics[width=0.6\linewidth]{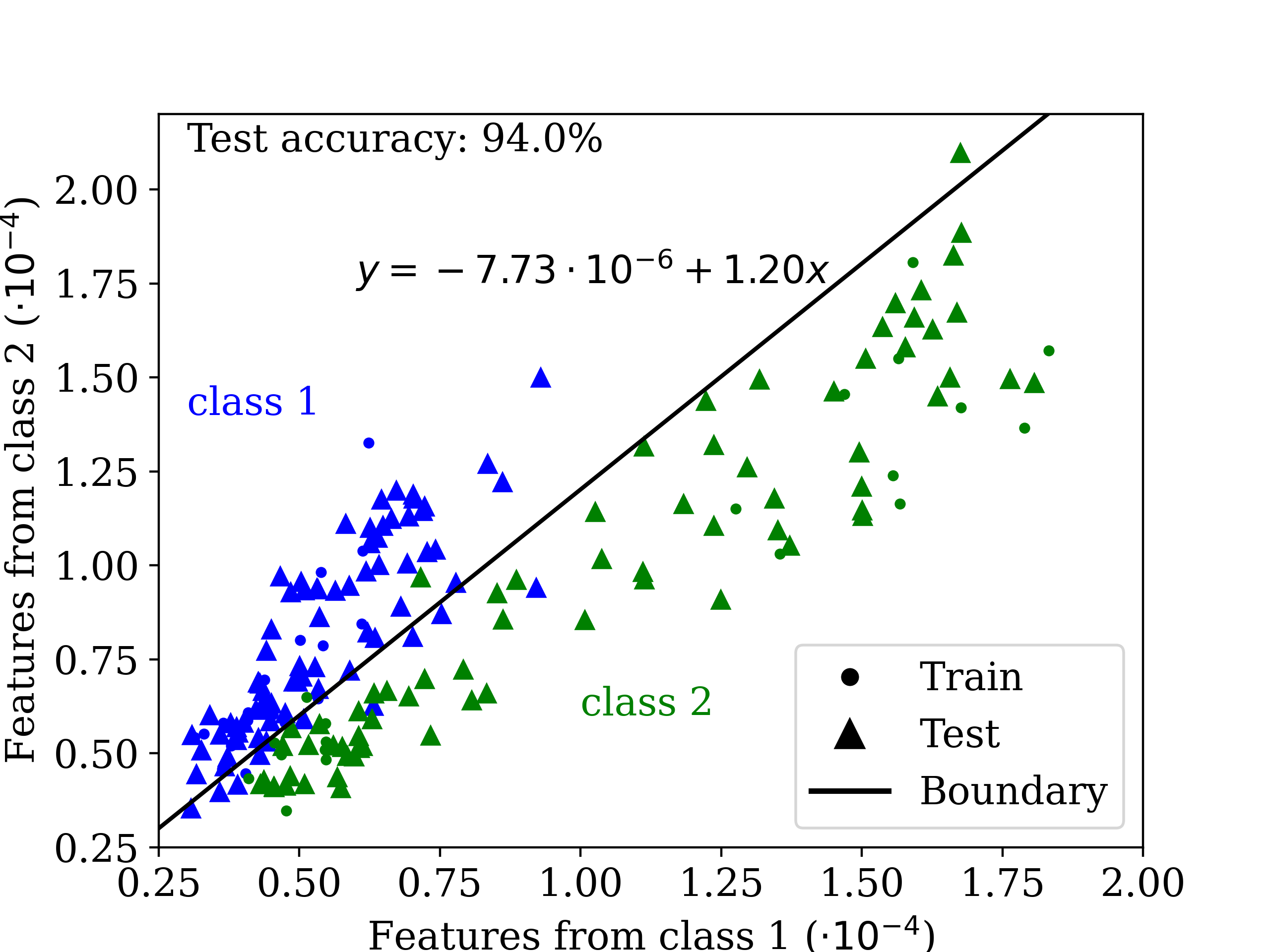}
    \caption{The transformation tries to make the data linearly separable. The \textit{GunPoint} dataset, \(r=25,\, l=4,\, k=1\), 1st sensor, mean with 5\(^\text{th}\) percentile as the feature extraction method. The instances corresponding to different classes are marked with different colors, the train instances are marked with dots, and the test instances with triangles. A linear separator (shown as a black line with its equation) is able to classify the data with 94.0\% test accuracy. Note: some of the correctly classified points are not plotted for better visibility.}
    \label{fig:GunPoint_plot}
\end{figure}

\noindent
The next example shows the transformed features, Figure \ref{fig:GunPoint_plot}, of the \textit{GunPoint} dataset from the UCR Time Series Classification Archive. The instances from class 1 are marked with blue and from class 2 with green color. It can be seen that the data corresponding to the same class tend to form linearly separable clusters. If the data are separated by the line shown in the plot, it results in 94\% classification accuracy. The train instances are marked with dots and the test instances are marked with triangles.

\subsection{The effect of changing the parameters}

\begin{figure}[h!]
    \centering
    \includegraphics[width=0.9\linewidth]{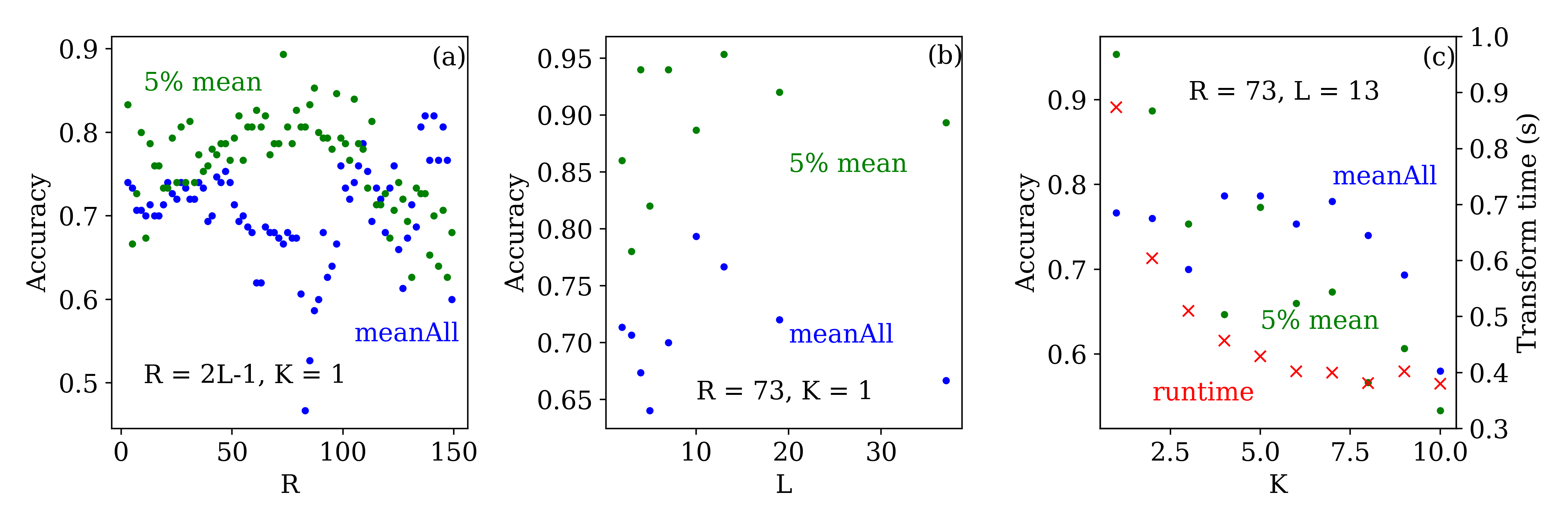}
    \caption{The effect of changing the parameters \(r, l\) and \(k\) on the \textit{GunPoint} database. The accuracies are calculated using linear discriminant on 2 features, similar to Figure \ref{fig:GunPoint_plot}. Parameter \(r\) describes the physical scale in which the shapelets are found, subfigure (a). The embedding dimension \(l\) describes the depth of the linear recursion, subfigure (b). Increasing \(k\) will lower the accuracy and the transformation time, subfigure (c). It depends on the database which \(r\) and \(l\) values separate the instances well and which feature extraction method should be used.}
    \label{fig:GunPoint_RLK_plot}
\end{figure}

\noindent
Next, the effect of changing the different parameters is discussed, the results are shown in Figure \ref{fig:GunPoint_RLK_plot}. The main parameters of the model are \(r, l\) and \(k\). In order to investigate their effect on the accuracy of the classification, the \textit{GunPoint} database is shown as an example. Then the accuracy is calculated by a linear separator always on 2 features, similar to Figure \ref{fig:GunPoint_plot}. In all the plots, 2 feature extraction methods are used: 5\(^\text{th}\) percentile then mean is marked with green, and the mean of all values is marked with blue. Also, in subplot (c) the runtimes are marked with red.

Parameter \(r\) describes the scale on which the patterns are found, subplot (a). It can be seen that most of the accuracies are around 75-85\% for the 5\% mean, and 60-80\% for the meanAll method. But there is one which is higher and one which is lower than the other accuracies. In that plot the \(l\) values can be calculated from the \(r\) values according to the relation \(r=2l-1\), where \(k=1\) is used.

Parameter \(l\) describes the embedding dimension, shown in subplot (b). Again, it can be seen that the 5\% mean gives higher accuracies than the meanAll method. In this plot \(r=73\) and \(k=1\) were used.

Parameter \(k\) describes the shift of the time window, subplot (c). It can be seen that increasing \(k\) will dramatically decrease the accuracy of the classification, but at the same time the runtime will be considerably reduced. Interestingly, the meanAll accuracies will stay at a constant level even for higher \(k\) values, while the 5\% mean values decrease quickly. An interesting case is \(k=2\) where the accuracy decreases only slightly but the runtime decreases by a third. So in some cases it might be beneficial to use \(k=2\) instead of \(k=1\) although generally \(k=1\) is a better choice. Increasing \(k\) further will decrease the runtime until it plateaus. 

\section{Conclusions}
\label{se:conclusions}

\noindent
This article described \texttt{ALT}, the Python package performing our novel method for univariate and multivariate TSC. The package provides an easy to use interface for the method, which has low computational cost and can be run on the GPU. In this paper we presented several examples demonstrating that the method tries and sometimes is able to create a linearly separable feature space, which can be classified with high accuracy using traditional classifiers such as KNN or SVM.

An earlier version of the method, LLT has already been demonstrated to be able to classify the \textit{Activity Recognition system based on Multisensor data fusion (AReM)} dataset~\cite{kurbucz2022facilitating} with high accuracy. It also improved the accuracy of traditional classifiers in predicting price movement of cryptocurrencies~\cite{kurbucz2023predicting}. More recently, it has been successfully applied to biomedical signal processing, providing a lightweight and transparent solution for ECG analysis in medical settings~\cite{posfay2024learning}.

Future research will focus on incorporating data-driven approaches to automatically tune the hyperparameters $(r, l, k)$, thereby minimizing the need for manual hyperparameter exploration. Additionally, we plan to investigate shapelet pruning strategies (as outlined in step [C] of Figure~\ref{fig:ALT}) to reduce computational overhead, enabling \texttt{ALT} to scale efficiently to large time series datasets without significant performance degradation. 
To enhance the interpretability of \texttt{ALT}, we aim to implement qualitative visualization techniques for the extracted shapelet vectors. Such visualizations could provide valuable insights into latent domain structures, making the transformation pipeline more transparent and easier to analyze. 
Lastly, we intend to evaluate ALT in specialized application areas such as multi-channel EEG monitoring and IoT anomaly detection. These domains offer unique challenges and opportunities for further performance improvements, while also emphasizing the importance of integrating domain-specific knowledge into the transformation process.

In conclusion, the value of the \texttt{ALT} package can be summarized as follows, it:
\begin{itemize}
\item Maps features into a linearly separable space using only a few hyperparameters.
\item Achieves state-of-the-art accuracy on benchmark TSC datasets.
\item Offers superior speed and transparency compared to existing methods.
\item Efficiently captures complex time series patterns across diverse temporal scales.
\end{itemize}

\bibliographystyle{elsarticle-num}
\bibliography{bibliography}

\begin{thebibliography}{10}
\expandafter\ifx\csname url\endcsname\relax
  \def\url#1{\texttt{#1}}\fi
\expandafter\ifx\csname urlprefix\endcsname\relax\def\urlprefix{URL }\fi
\expandafter\ifx\csname href\endcsname\relax
  \def\href#1#2{#2} \def\path#1{#1}\fi

\bibitem{fulcher2014highly}
B.~D. Fulcher, N.~S. Jones, Highly comparative feature-based time-series classification, IEEE Transactions on Knowledge and Data Engineering 26~(12) (2014) 3026--3037.

\bibitem{senin2008dynamic}
P.~Senin, Dynamic time warping algorithm review, Information and Computer Science Department University of Hawaii at Manoa Honolulu, USA 855~(1-23) (2008) 40.

\bibitem{fawaz2019deep}
H.~I. Fawaz, G.~Forestier, J.~Weber, L.~Idoumghar, P.-A. Muller, Deep learning for time series classification: a review, Data Mining and Knowledge Discovery 33~(4) (2019) 917--963.

\bibitem{agrawal1993efficient}
R.~Agrawal, C.~Faloutsos, A.~N. Swami, Efficient similarity search in sequence databases, in: Proceedings of the 4th International Conference on Foundations of Data Organization and Algorithms (FODO), Springer, 1993, pp. 69--84.

\bibitem{box2015time}
G.~E.~P. Box, G.~M. Jenkins, G.~C. Reinsel, G.~M. Ljung, Time series analysis: forecasting and control, John Wiley \& Sons, 2015.

\bibitem{ye2009time}
L.~Ye, E.~Keogh, Time series shapelets: a new primitive for data mining, in: Proceedings of the 15th ACM SIGKDD international conference on Knowledge discovery and data mining, ACM, 2009, pp. 947--956.

\bibitem{karim2019multivariate}
F.~Karim, S.~Majumdar, H.~Darabi, S.~Chen, Multivariate lstm-fcns for time series classification, Neural Networks 116 (2019) 237--245.

\bibitem{zheng2014time}
Y.~Zheng, Q.~Liu, E.~Chen, Y.~Ge, J.~Zhao, Time series classification using multi-channels deep convolutional neural networks, in: International Conference on Web-Age Information Management, Springer, 2014, pp. 298--310.

\bibitem{jakovac2022reconstruction}
A.~Jakov{\'a}c, M.~T. Kurbucz, P.~P{\'o}sfay, Reconstruction of observed mechanical motions with artificial intelligence tools, New Journal of Physics 24~(7) (2022) 073021.

\bibitem{jakovac2021time}
A.~Jakovac, Time series analysis with dynamic law exploration, arXiv preprint arXiv:2104.10970 (2021).

\bibitem{kurbucz2022facilitating}
M.~T. Kurbucz, P.~P{\'o}sfay, A.~Jakov{\'a}c, Facilitating time series classification by linear law-based feature space transformation, Scientific Reports 12~(1) (2022) 18026.

\bibitem{KURBUCZ2024101623}
M.~T. Kurbucz, P.~Pósfay, A.~Jakovác, \href{https://www.sciencedirect.com/science/article/pii/S2352711023003199}{{LLT}: An r package for linear law-based feature space transformation}, SoftwareX 25 (2024) 101623.
\newblock \href {https://doi.org/https://doi.org/10.1016/j.softx.2023.101623} {\path{doi:https://doi.org/10.1016/j.softx.2023.101623}}.
\newline\urlprefix\url{https://www.sciencedirect.com/science/article/pii/S2352711023003199}

\bibitem{kurbucz2025adaptive}
M.~T. Kurbucz, B.~Haj{\'o}s, B.~P. Halmos, V.~{\'A}. Moln{\'a}r, A.~Jakov{\'a}c, Adaptive law-based transformation ({ALT}): A lightweight feature representation for time series classification, arXiv preprint arXiv:2501.09217 (2025).

\bibitem{wu2024invariant}
Y.~Wu, Y.~Yang, J.~S. Xiao, C.~Zhou, H.~Sui, H.~Li, \href{https://openreview.net/forum?id=Ex6wAivo7G}{Invariant spatiotemporal representation learning for cross-patient seizure classification}, in: The First Workshop on NeuroAI @ NeurIPS2024, 2024.
\newline\urlprefix\url{https://openreview.net/forum?id=Ex6wAivo7G}

\bibitem{kurbucz2023predicting}
M.~T. Kurbucz, P.~P{\'o}sfay, A.~Jakov{\'a}c, Predicting the price movement of cryptocurrencies using linear law-based transformation, arXiv preprint arXiv:2305.04884 (2023).

\bibitem{takens1981dynamical}
F.~Takens, Dynamical systems and turbulence, Warwick, 1980 (1981) 366--381.

\bibitem{gao2021human}
F.~Gao, T.~Tian, T.~Yao, Q.~Zhang, Human gait recognition based on multiple feature combination and parameter optimization algorithms, Computational Intelligence and Neuroscience 2021~(1) (2021) 6693206.

\bibitem{UCRArchive2018}
H.~A. Dau, E.~Keogh, K.~Kamgar, C.-C.~M. Yeh, Y.~Zhu, S.~Gharghabi, C.~A. Ratanamahatana, Yanping, B.~Hu, N.~Begum, A.~Bagnall, A.~Mueen, G.~Batista, Hexagon-ML, The ucr time series classification archive, \url{https://www.cs.ucr.edu/~eamonn/time_series_data_2018/} (October 2018).

\bibitem{ratanamahatana2005three}
C.~A. Ratanamahatana, E.~Keogh, Three myths about dynamic time warping data mining, in: Proceedings of the 2005 SIAM international conference on data mining, SIAM, 2005, pp. 506--510.

\bibitem{posfay2024learning}
P.~Pósfay, M.~T. Kurbucz, P.~Kovács, A.~Jakovác, \href{https://doi.org/10.48550/arXiv.2307.01930}{Learning {ECG} signal features without backpropagation using linear laws}, arXiv preprint arXiv:2307.01930v2Version 2, last revised 20 Dec 2024 (2024).
\newline\urlprefix\url{https://doi.org/10.48550/arXiv.2307.01930}

\end{thebibliography}

\section*{Acknowledgments}

\noindent
The research was supported by the Hungarian Government and the European Union in the framework of a Grant Agreement No. MILAB RRF-2.3.1-21-2022-00004. Project no. PD142593 was implemented with the support provided by the Ministry of Culture and Innovation of Hungary from the National Research, Development, and Innovation Fund, financed under the PD\_22 ``OTKA'' funding scheme.
\end{document}